\documentclass[10pt,twocolumn,letterpaper]{article}

\usepackage[pagenumbers]{iccv} 

\usepackage{times}
\usepackage{epsfig}
\usepackage{graphicx}
\usepackage{amsmath}
\usepackage{amssymb}
\usepackage{booktabs}
\usepackage{color}
\usepackage{caption}
\usepackage{subcaption}
\usepackage{makecell}
\usepackage{multirow}

\usepackage{bbding}
\usepackage{url}

\usepackage{float}
\usepackage{balance}

\usepackage{tcolorbox}
\usepackage{xcolor}
\usepackage{circledsteps}

\definecolor{iccvblue}{rgb}{0.21,0.49,0.74}
\definecolor{salmon}{RGB}{255,103,125}
\usepackage[pagebackref,breaklinks,colorlinks,allcolors=iccvblue]{hyperref}

\def\paperID{306} 
\def\confName{ICCV}
\def\confYear{2025}

\title{Benchmarking Multimodal CoT Reward Model \underline{S}tepwise by \underline{Vi}sual \underline{P}rogram}

\author{
Minghe Gao\textsuperscript{1}, Xuqi Liu\textsuperscript{1}, Zhongqi Yue\textsuperscript{2}, Yang Wu\textsuperscript{3}, Shuang Chen\textsuperscript{1}\\ 
Juncheng Li\textsuperscript{1}\dag, Siliang Tang\textsuperscript{1}, Fei Wu\textsuperscript{1}, Tat-Seng Chua\textsuperscript{4}, Yueting Zhuang\textsuperscript{1}\\
\textsuperscript{1}Zhejiang University
\textsuperscript{2}Nanyang Technological University\\
\textsuperscript{3}Ant Group
\textsuperscript{4}National University of Singapore \dag \textit{Corresponding Author}
}

\begin{document}
\maketitle

\begin{abstract}
\vspace{-4mm}

Recent advancements in reward signal usage for Large Language Models (LLMs) are remarkable. However, significant challenges exist when transitioning reward signal to the multimodal domain, including labor-intensive annotations, over-reliance on one-step rewards, and inadequate evaluation. To address these issues, we propose \texttt{\textit{SVIP}}, a novel approach to train a step-level multi-dimensional Chain-of-Thought~(CoT) reward model automatically. It generates code for solving visual tasks and transforms the analysis of code blocks into the evaluation of CoT step as training samples. Then, we train \texttt{\textit{SVIP-Reward}} model using a multi-head attention mechanism called TriAtt-CoT. The advantages of \texttt{\textit{SVIP-Reward}} are evident throughout the entire process of MLLM. We also introduce a benchmark for CoT reward model training and testing. Experimental results demonstrate that \texttt{\textit{SVIP-Reward}} improves MLLM performance across training and inference-time scaling, yielding better results on benchmarks while reducing hallucinations and enhancing reasoning ability. The code and data are open-sourced at: \url{https://github.com/minghehe-nobug/SVIP}

\vspace{-4mm}
\end{abstract}

\section{Introduction}

Recent research indicates that rather than allocating extra computational resources to pre-training, it is more effective to utilize reward signals to enhance LLM reasoning performance during test/inference-time scaling or post-training phases~\cite{lightman2023letsverifystepstep,snell2024scalingllmtesttimecompute,xin2024deepseekproverv15harnessingproofassistant,wang2024mathshepherdverifyreinforcellms,uesato2022solvingmathwordproblems}. 
For instance, OpenAI’s o1~\cite{openaio1} is the first to introduce inference-time scaling that selects the most promising candidate during inference based on the reward model's score.
Deepseek-R1~\cite{deepseekai2025deepseekr1incentivizingreasoningcapability} adopts a rule-based reward system to facilitate reinforcement learning. 
Given the critical role of the reward signals in improving reasoning capabilities, extending this advantageous mechanism to the Multimodal Large Language Models (MLLMs) domain could similarly enhance their performance.

\begin{figure}
  \centering
    \includegraphics[width=0.8\linewidth]{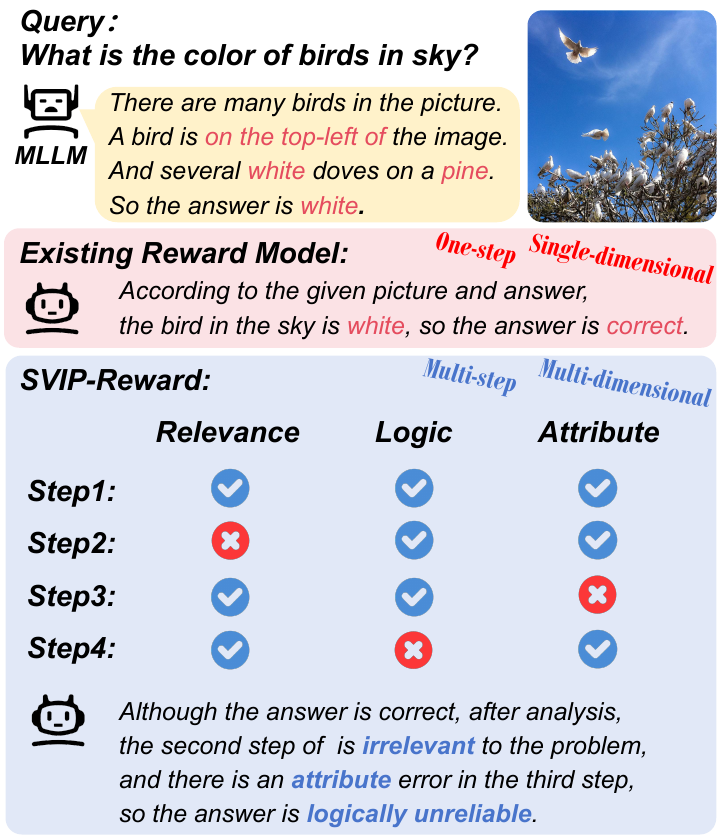}
    \vspace{-3mm}
    \caption{Existing reward models cannot provide multi-dimensional assessments at the step level, while \texttt{\textit{SVIP}} evaluates each CoT step across three dimensions.}
    \label{fig_1}
    \vspace{-5mm}
\end{figure}

\begin{figure*}
  \centering
    \includegraphics[width=\linewidth]{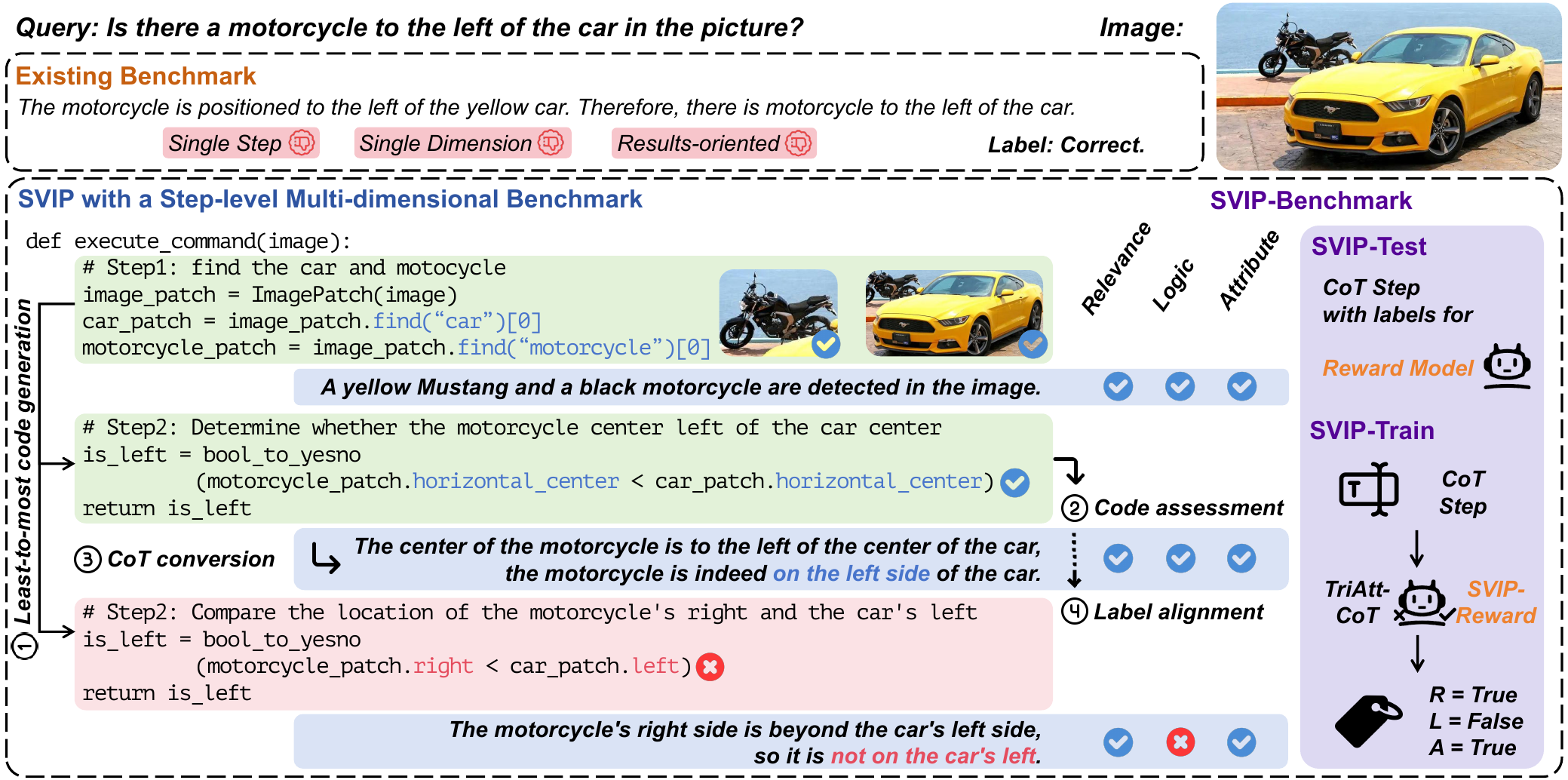}
    \vspace{-7mm}
    \caption{The framework of \texttt{\textit{SVIP}}: First, we generate multiple candidate code blocks for a VQA pair by the least-to-most prompt. Then, these code blocks are evaluated based on their compilability, logic, and function calls and align with the converted CoT step labels. Finally, these step-level, multi-dimensional data are used to train \texttt{\textit{SVIP-Reward}}.}
    \label{fig_3}
    \vspace{-3mm}
\end{figure*}

However, compared to established practices in LLMs, implementing reward signals in MLLMs has challenges:
\textbf{1)~\textcolor{salmon}{Labor-intensive and non-scalable annotation}}: Manually generating CoT, along with human evaluations or template-based assessment~\cite{wei2023chainofthoughtpromptingelicitsreasoning} for reward signals are unfeasible due to their high labor intensity and lack of scalability, particularly more costly in the multimodal domain.
\textbf{2)~\textcolor{salmon}{Over reliance on one-step reward}}: Models like R1~\cite{deepseekai2025deepseekr1incentivizingreasoningcapability} \newline mainly target tasks (e.g., mathematical or coding problems) with a single, unique outcome, using the final result as the sole output reward signal. This narrowly defined approach neglects supervision for intermediate steps.
\textbf{3)~\textcolor{salmon}{Single-dimensional and inadequate evaluation}}: Current CoT evaluation scores from reward models are one-dimensional~\cite{wijaya2024multimodalpreferencedatasynthetic,li2024vlrewardbenchchallengingbenchmarkvisionlanguage,zang2025internlmxcomposer25rewardsimpleeffectivemultimodal}. Such frameworks risk overfitting to correctness or preference and neglecting other critical dimensions in tasks that require more than mere accuracy.
\textbf{Collectively}, these issues constrain existing reward signal to a “\textit{\textcolor{gray}{non-automatic, results-oriented, single-dimensional}}” paradigm as depicted in Figure~\ref{fig_1}, failing to capture complex, step-by-step reasoning. Ideally, as shown in the bottom section of Figure~\ref{fig_1}, a CoT reward model can \textbf{automatically provide a step-level multi-dimensional signal}, enhancing both the training and inference phases of MLLMs.


Encouragingly, we are proud to be the first in discovering that Visual Programming~\cite{Gupta_2023_CVPR,Suris_2023_ICCV} can adeptly address the aforementioned challenges in the multimodal domain:
\textbf{1)~\textcolor{iccvblue}{Automated reasoning chain generation}}: Visual Programming is a burgeoning field that uses a code pre-trained model to generate Python programs for vision tasks in a zero-shot manner. It allows for extensive code generation with the minimal cost and without the need for supervision. \newline
\textbf{2)~\textcolor{iccvblue}{Step-level task decomposition}}: Visual Programming inherently organizes logical steps into a program flow symbolically, closely mirroring the CoT approach where complex tasks are decomposed into sequential segments. This alignment with CoT step-by-step nature ensures each code block in Visual Programming correlates with a CoT reasoning step, facilitating a seamless transition from code to CoT. \newline
\textbf{3)~\textcolor{iccvblue}{Multi-dimensional statement evaluation}}: Furthermore, the modular design of Visual Programming supports multi-dimensional evaluation. Each module can be independently assessed through intermediate variables, code logic, and compilation data, reflecting key aspects such as data integrity and logical flow. This provides a more comprehensive assessment of the reasoning process.


Inspired by Visual Programming, we introduce \textit{\texttt{SVIP}}, a method that generates executable code to solve visual tasks and seamlessly translates the creation of code and its analytical evaluation into corresponding steps and assessments within the CoT framework. As shown in Figure~\ref{fig_3}, this is a two-phase code-to-CoT process. In the \textbf{code generation~(\Circled{1})~ and assessment~(\Circled{2}) phase}, we begin by generating Python programs for each task, using least-to-most prompting. Subsequently, we conduct a multi-dimensional analysis of these blocks by evaluating their compilability, logic, and function calls. In the \textbf{CoT conversion~(\Circled{3})~ and alignment~(\Circled{4}) phase}, we convert the generated code into sequential CoT statements and then map the detailed analysis of each code block into step-level multi-dimensional annotations within these statements. \textbf{Finally}, to capitalize on the evaluations derived from \textit{\texttt{SVIP}}, we train our reward model \texttt{\textit{SVIP-Reward}} using a multi-head attention mechanism called \textit{\textbf{TriAtt-CoT}}. This mechanism is specifically designed to attentively weigh and integrate the varied dimensions of relevance, logic, and attributes assessed in each CoT step.

\begin{figure*}[!t]
    \hfill
    \begin{minipage}{0.7\textwidth}
        \centering
        \includegraphics[width=\textwidth]
        {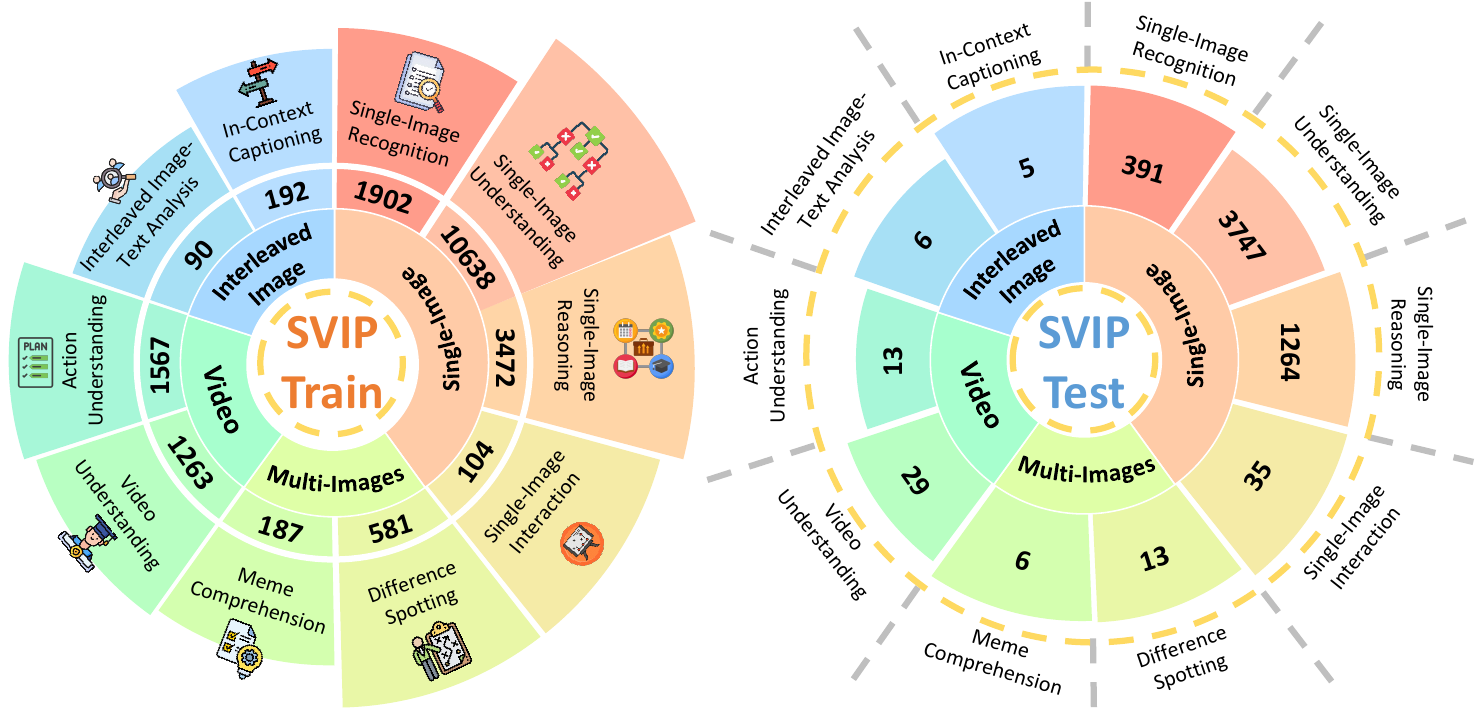}  
        \caption{Data distribution of \textit{SVIP-Train} and \textit{SVIP-Test}}
        \label{fig_2}
    \end{minipage}
    \begin{minipage}{0.29\textwidth}
        \resizebox{1.0\textwidth}{!}{
        \renewcommand{\arraystretch}{1}
            \begin{tabular}{@{ } l | c  c  c  c  c }
            \makecell[l]{Reward\\Model\\Benchmark} & \rotatebox{90}{Modality} & \rotatebox{90}{Multi-dimensional} & \rotatebox{90}{Step-level} & \rotatebox{90}{Interleaved} & \rotatebox{90}{Scalable} \\
            \midrule
            \makecell[l]{RMB\\~\cite{zhou2024rmb}} & \textit{Textual} & \Checkmark & \XSolidBrush & \XSolidBrush & \XSolidBrush \\
            \makecell[l]{PRMBench\\~\cite{song2025prmbench}} & \textit{Textual} & \Checkmark & \Checkmark & \XSolidBrush & \XSolidBrush \\
            \midrule
            \makecell[l]{MJ-Bench\\~\cite{chen2024mjbenchmultimodalrewardmodel}} & \makecell{\textit{Textual}\\\textit{Visual}} & \XSolidBrush & \XSolidBrush & \XSolidBrush & \XSolidBrush\\
            \makecell[l]{VL-Reward\\Bench~\cite{li2024vlrewardbench}} & \makecell{\textit{Textual}\\\textit{Visual}} & \XSolidBrush & \XSolidBrush & \XSolidBrush & \Checkmark\\
            \midrule
            \textit{SVIP-Test} & \makecell{\textit{Textual}\\\textit{Visual}} & \Checkmark & \Checkmark & \Checkmark & \Checkmark\\
            \midrule
            \end{tabular}
        }
        \vspace{-3mm}
        \captionof{table}{Benchmark comparison.}\label{table_1}
    \end{minipage}
    \vspace{-5mm}
\end{figure*}

The advantages of \texttt{\textit{SVIP-Reward}} are apparent throughout the entire MLLM pipeline. \textbf{During the data cleaning phase}, \texttt{\textit{SVIP-Reward}} enhances the selection process by identifying high-quality multimodal reasoning samples based on critical dimensions. \textbf{During the training phase}, it automatically provides multi-dimensional supervisory signals at each step for reinforcement learning. \textbf{During the inference phase}, it leverages reward signals to select the best candidate step for \textbf{test-time scaling}, thereby enhancing the model’s reasoning ability at a relatively low computational cost ~\cite{deepseekai2025deepseekr1incentivizingreasoningcapability,zang2025internlmxcomposer25rewardsimpleeffectivemultimodal}.

To further promote the in-depth study of the reward model, and during \texttt{\textit{SVIP-Reward}} developing, we introduce a two-part step-level multi-dimensional CoT benchmark (Figure~\ref{fig_2}): \textit{SVIP-Train}, containing 7,948 program-derived CoT with 20,000 steps for reward model training, and \textit{SVIP-Test}, curated by professionals, containing 1,934 program-derived CoT with 5,509 steps for evaluation. This scalable dataset spans 24 tasks and supports single-/multi-image and video modalities. In our subsequent analysis of human preferences, the samples annotated by \textit{\texttt{SVIP}} showed a high degree of overlap with those selected by humans. To the best of our knowledge, and in contrast to other existing reward model benchmarks (Table~\ref{table_1}), our new benchmark is the only one in the MLLM field that can provide precise step-level multi-dimensional CoT annotations.

Through extensive experiments, we demonstrate: 
1) Effectiveness of step-level, multi-dimensional data: Both Qwen2-VL-7B~\cite{wang2024qwen2vlenhancingvisionlanguagemodels} and InternVL2.5-2B~\cite{chen2024expanding} tuned with \textit{SVIP-Train} under \textit{Tuning} setting achieve 6.3\% and 2.3\% improvement, respectively, over the baseline on \textit{SVIP-Test}.
2) Synergistic gain from \textit{\textbf{TriAtt-CoT}}: \textit{\textbf{TriAtt-CoT}} effectively distinguishes the three-dimensional labels. The \textit{\texttt{SVIP-Reward}} models demonstrate an average improvement of 5.95\% on the \textit{SVIP-Test} compared to \textit{Tuning} setting, highlighting the synergistic benefits of it.
3) Valuable fine-grained reward by \textit{\texttt{SVIP-Reward}} for training: MLLMs with supervisory signals during training perform better on, MME~\cite{fu2024mmecomprehensiveevaluationbenchmark}, MMMU~\cite{yue2024mmmu} etc., known for emphasis on reasoning.
4) Effective guidance in inference: Integrating \textit{\texttt{SVIP-Reward}} with sampling-based algorithms for test-time scaling further improves MLLMs’ reasoning accuracy.
Overall, these enhancements also lead to a notable reduction in hallucinations~\cite{li2023evaluating}.
In summary, our contributions are as follows:

\begin{itemize}

    \item{We revisit the evaluation of CoT as an analysis of the program, defining three dimensions for step-level CoT evaluation and introducing a least-to-most sampling method for collecting fine-grained reward signals.}

    \item{We propose \textbf{\textit{TriAtt-CoT}} to train step-level, multi-dimensional CoT reward models. Compared to direct training with ground-truth answers, it demonstrates superior alignment and performance.}

    \item{Experiments show that \textit{\texttt{SVIP-Reward}} achieves SOTA performance on \textit{SVIP-Test} and enhances the MLLMs' reasoning capabilities in both training and inference, while significantly reducing hallucinations.}

\end{itemize}

\section{Related Works}

\subsection{Multimodal CoT Fine-Tuning}

Multimodal CoT fine-tuning is a process designed to enhance the reasoning capabilities of large multimodal models by integrating information from both textual and visual modalities~\cite{NEURIPS2024_0ff38d72,zhang2024cocotcontrastivechainofthoughtprompting,chen2024m}. This approach trains models to perform step-by-step reasoning, improving their ability to focus on critical regions within visual inputs and make connections between text and images. Through fine-tuning, models are guided to reason across multiple domains, steps, and inputs, addressing challenges such as fine-grained visual perception and accurate multi-image comprehension.

Despite the recognition of the importance of CoT steps and their equivalence to the final result in many works, CoT is still treated as a whole in practice rather than at the step level~\cite{hsieh2023distilling}. This results in an inability to evaluate each individual step within the CoT. In complex real-world tasks, the CoT for the same problem should not be unique~\cite{yao2024tree}. The existing methods limit the model’s ability to learn how to solve a complex problem step-by-step and transfer it to similar problems. In contrast, \texttt{\textit{SVIP}} can train reward models using fine-grained step-level labels step by step.

\subsection{Visual Programming as Decision Sequence}

Visual Programming ~\cite{Gupta_2023_CVPR, Suris_2023_ICCV} is an emerging field that utilizes neural symbols or Python modules for task synthesis and execution. Its advantage lies in the ability to call visual modules through code, with the reasoning logic of the program explicitly presented in the form of code ~\cite{NEURIPS2023_871ed095}. Recently, several works have attempted to leverage the interpretability of visual programming, such as VPD ~\cite{hu2024visual}, which simplifies multimodal learning by distilling tool use and programmatic reasoning into smaller models and De-fine~\cite{gao2024fine} reimagines visual programming as a task of modular programming and optimization through feedback.

Inspired by these works, we further considered treating the execution process of code and its intermediate variables
as a CoT decision sequence. The advantage of using programs as CoT in SVIP is evident: code is inherently verifiable and evaluable. Fact~\cite{gao2024fact} eliminates redundant elements through static analysis of Python code’s AST and converts them into natural language CoT, while PropTest~\cite{koo2024proptest} generates test code to validate the answers. These works provide a solid backbone for us—we can transform the evaluation of CoT steps into the evaluation of code, thereby addressing the challenge of evaluating CoT steps in natural language.

\section{Method}
\label{sec3}

We introduce \texttt{\textit{SVIP}}, a step-level CoT generation and multi-dimensional labeling framework for reward model training, shown in Figure ~\ref{fig_3}. Given a candidate visual task, we initially generate several candidate codes by least-to-most prompting (Section~\ref{sec3.1}). These codes are then executed block by block to obtain intermediate results and provide a comprehensive evaluation (Section~\ref{sec3.2}). Next, these code analyses will be converted into annotations for the CoT step accordingly (Section~\ref{sec3.3}). Finally, we propose \textit{\textbf{TriAtt-CoT}} to train the \textit{\texttt{SVIP-Reward}} model. (Section~\ref{sec3.4}).

\subsection{Least-to-Most Visual Program Generation}
\label{sec3.1}

Visual programming is typically a one-step process, generating code in a single pass from beginning to end. However, even with varied outputs via temperature adjustment, it lacks iterative interaction and alternative path exploration. In \texttt{\textit{SVIP}}, we adopt a least-to-most~\cite{zhou2022least} prompt strategy, incrementally generating code starting with the current block. This approach allows for the generation of subsequent code blocks, enabling multiple decision-making pathways and enhancing dynamic exploration of the visual task.

Specifically, we input a query $q$, visual input $v$, and instruction prompt $p$ (as detailed in Appendix D) to a code generator $\pi$. The goal of the code generator is not to generate a complete executable code but rather to generate a code block $b_1=\pi(q,v,p)$ that completes the first subtask. Based on the generated code sequence $c= \left[ b_1,...,b_n \right] $, we then generate the next step $b_{n+1} = \pi(q,v,p,c)$, continuing this process until a termination character is reached, at this time $c$ represents a complete program. Through this process, we can generate the visual program block by block. Furthermore, at each step, we generate $X$ subnodes, which leads to a tree structure $T=\{b_1,...,b_n\}$ with code blocks $b$ as the nodes. The path from the root to the leaf nodes represents a complete program, and we use $C=\{c_1,...,c_k\}$ to denote the set of codes that solve the problem $q$. The least-to-most visual program generation in \texttt{\textit{SVIP}} scales adaptively with the query's complexity. More complex problems generate more subtasks and step-level samples, reflecting the intuition that harder problems warrant more solutions.

\subsection{Code Execution and Block Labeling}
\label{sec3.2}

Code blocks serve as fine-grained outputs meeting step-level granularity requirements. However, without intermediate variables from execution, evaluating the logic and answer in a code block is impossible. Therefore, generating the code set $C$, we apply an execution engine $\phi$ to execute the program on the visual input, obtaining intermediate variables $T_k = \phi(c_k,v) $ where $k \in \{1,...,K\}$. We then extract these temporary variables $T_k=\{t_1,t_2,...,t_k\}$ and match them with the corresponding nodes, where $t_m$ represents the intermediate variable (e.g., image patches, strings, lists) during the execution of code block $b_m$.

Thanks to the code-independent statements and the rigorous dependencies between variables, we are able to analyze and evaluate the blocks from three distinct dimensions:

\textbf{Relevance}: For code blocks that fail to compile or encounter errors during compilation, \texttt{\textit{SVIP}} will not be able to obtain intermediate variables, and subsequent code blocks will also fail to execute. In such cases, the program will objectively present the error messages from the Integrated Development Environment (IDE), resulting in feedback that is unrelated to the tasks or visual input. We classify such compilation failures as negative samples, while successful compilations, which provide essential intermediate information for reasoning, are considered positive samples.

\textbf{Logic}: Visual programming can explicitly represent natural language abstractions' logical relationships using logical symbols. Additionally, mature and convenient methods are available to verify the correctness of these logical relationships by generating test cases or checking variable formats. Therefore, we use an enhanced version of PropTest~\cite{koo2024proptest} to validate code blocks with logical symbols. PropTest performs format checks on intermediate variables, and we manually define input-output test cases for basic logical operations (such as AND, OR, NOT, etc.). Ultimately, code blocks that pass PropTest validation are considered logically correct.

\begin{figure}
  \centering
    \includegraphics[width=0.95\linewidth]{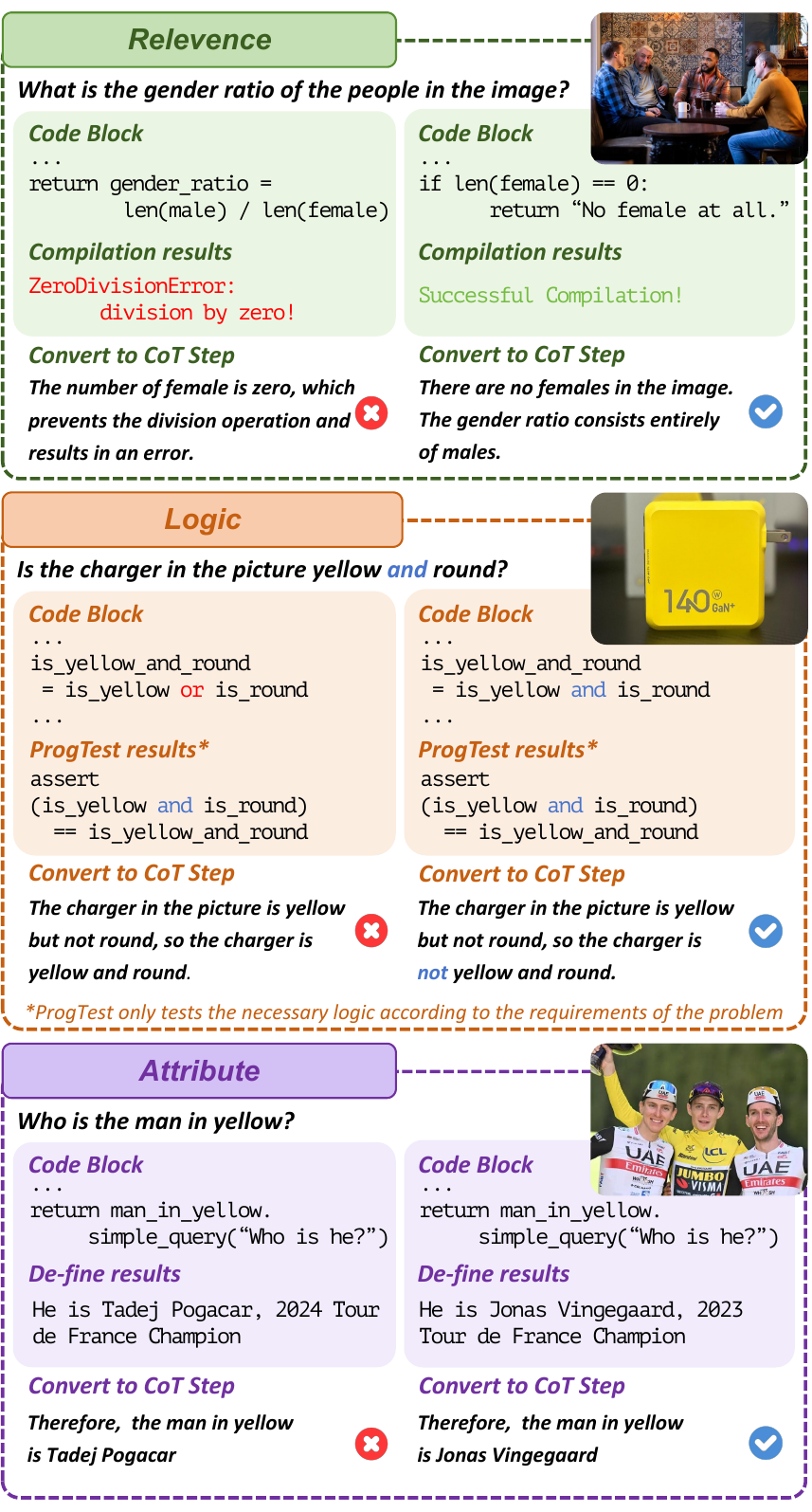}
    \vspace{-3mm}
    \caption{Three CoT step labels defined by \texttt{\textit{SVIP}} and the corresponding assessment through program analyze.}
    \label{fig_4}
    \vspace{-5mm}
\end{figure}

\textbf{Attribute}: Visual programming may face errors when calling external functions due to module limitations. These errors do not stem from decision-making failures in the visual programming process, but rather from functional errors, such as mistakenly recognizing ``Labrador'' as ``dog''. Such attribute errors, triggered by external knowledge, can be detected through cross-validation with other models. Therefore, we employ the feedback from De-fine~\cite{gao2024fine} to validate the return values of invoked module functions. Labels that pass the cross-validation process are defined as attribute-correct. For ease of understanding, we show examples for each of the three labels in Figure~\ref{fig_4}.

\subsection{CoT Step Conversion}
\label{sec3.3}

Previous works such as VPD~\cite{hu2024visual} and Fact~\cite{gao2024fact} have demonstrated the correspondence between code execution trajectories and multimodal CoT. Therefore, we use an MLLM as the CoT Generator $G$ to synthesize a new multimodal CoT step $s=G(q,v,b_m,t_m)$ from the code block and its corresponding intermediate variables. It is important to note that in $t_m$, we represent image regions as textual descriptions using bounding boxes, and all intermediate variables are stored in a dictionary structure. As a result, $s$ becomes an interleaved multimodal CoT step. The final output is the set of CoT steps $S=\{s_1,...,s_M\}$. Under the constraints of MLLM, multimodal CoT steps faithfully represent the execution trajectory of the code. This means that if a code block encounters a logical or factual error during execution, it will be reflected in the CoT step as well. Each CoT step is associated with labels from three distinct dimensions, as training samples to train \texttt{\textit{SVIP-Reward}}.

\subsection{\textbf{\textit{TriAtt-CoT}} Reward Model Training}
\label{sec3.4}

To effectively manage multiple objectives within a task and avoid potential conflicts that could skew the model's focus, we introduce \textbf{\textit{TriAtt-CoT}} to train our \texttt{\textit{SVIP-Reward}} incorporates a tri-head attention layer. This initial layer is critical for accurately extracting input features from diverse data sources, ensuring a balanced consideration of competing objectives during the model's decision-making process.

Following feature extraction, \textbf{\textit{TriAtt-CoT}} employs these features to compute a contrastive loss against the labels. This method helps mitigate objective conflicts by promoting a balance, thereby refining the model's decision-making capabilities. The contrastive loss serves as an integral part of the model's optimization strategy, fostering a deeper integration of the multiple dimensions assessed.

The final enhancement in our approach involves replacing the base model’s traditional decoder layer with a scoring head. This scoring head predicts the reward scores by evaluating all dimensions of the task processed by the model, leading to step-level multi-dimensional reward calculations.

\begin{table*}
\centering 
\vspace{-3mm}
\caption{Result on \textit{SVIP-Test} and other CoT benchmarks.}
\vspace{-3mm}
\label{table2}
\resizebox{\textwidth}{!}{
\begin{tabular}{l | c c c c c c c} 
\midrule
   &   &   & \multicolumn{5}{c}{\textit{SVIP-Test}} \\
 Reward Model & MJ-Bench & VLRewardBench & Overall & \multicolumn{4}{c}{Step}\\
  \cmidrule(lr){4-4}  \cmidrule(lr){5-8}
 & & & Correctness & Relevance & Logic & Attribute & Avg \\
\midrule
GPT-4o & 65.8 & 65.8 & 58.4 & 82.2 & 54.7 & 56.8 & 64.6\\
IXC-2.5-Reward-7B &  69.2 & 65.8 & 63.7 & 89.4 & 60.1 & 53.6 & 67.7\\
\midrule
Qwen2-VL-7B (\textit{Zero-shot}) & 72.3 & 28.3 & 55.1 & 73.8 & 50.2 & 47.3 & 57.1 \\ 
Qwen2-VL-7B (\textit{Tuning}) & 73.1 & 55.4 & 63.9 & 81.5 & 55.9 & 52.7 & 63.4\\ 
Qwen2-VL-7B (\texttt{\textit{SVIP-Reward}}) & \textbf{73.7} & \textbf{68.0} & \textbf{67.3} & \textbf{92.3} & \textbf{61.0} & \textbf{58.9} & \textbf{70.7}\\
\midrule
InternVL-2.5-2B (\textit{Zero-shot}) & 48.0 & 35.1 & 46.2 & 75.6 & 52.1 & 51.1 & 59.6\\
InternVL-2.5-2B (\textit{Tuning}) & 50.4 & 49.6 & 54.5 & 78.6 & 53.5 & 53.5 & 61.9 \\ 
InternVL-2.5-2B (\texttt{\textit{SVIP-Reward}}) & \textbf{51.6} & \textbf{56.3} & \textbf{60.8} & \textbf{87.0} & \textbf{56.4} & \textbf{56.2} & \textbf{66.5}\\
\midrule
\end{tabular}
}
\vspace{-5mm}
\end{table*}

\section{\texttt{\textit{SVIP}} Benchmark}

Building on the automated step-level multi-dimensional CoT annotation outlined in the methods section, we create a CoT step benchmark including \textit{SVIP-Train} and \textit{SVIP-Test} with dimensions of relevance, logic, and attribute. 

\textbf{Dataset collection and construction.} \textit{SVIP-Train} is originated from SeedBench2~\cite{Li_2024_CVPR} and includes tasks with single images, multiple images, and videos. We expanded this dataset by generating 20,000 steps from 7,948 samples, crafting a scalable, automated training dataset that requires no manual annotation and is adaptable to any visual task. Future enhancements could exploit the difficulty levels in \textit{SVIP-Train} to promote an easy-to-hard generalization strategy. To further the development of multimodal CoT reward models, we meticulously selected 5509 CoT steps that do not overlap with \textit{SVIP-Train} to form the \textit{SVIP-Test} benchmark. To our knowledge, this benchmark is the only one in the MLLM field, offering step-level multi-dimensional testing for reward model performance. We compiled an analysis (Figure~\ref{fig_2}) for \textit{SVIP-Train} and \textit{SVIP-Test} to illustrate data distributions and conducted a comparison of \textit{SVIP-Test} with other benchmarks (Table~\ref{table_1}).

\textbf{Human acceptance analysis.} Simultaneously with selection, we also analyzed the filtered data for human acceptance. The results revealed that data annotated by the SVIP method exhibited an 83\% similarity with manually annotated data, demonstrating that this step-level, multi-dimensional annotation approach can be robustly extended to other datasets while maintaining high credibility. We have also included examples of correct and failure case in the Appendix A.

\textbf{Task definition and evaluation metric.} We divide the tasks into two categories: overall testing and step-level testing. The former evaluates the model's responses solely based on correctness, with labels directly inherited from SeedBench2~\cite{Li_2024_CVPR}. The latter assesses the reward signals across three dimensions to evaluate consistency with the labels. To facilitate a user-friendly evaluation applicable to most reward models, we adopt accuracy as the criterion in our evaluation metrics, simplifying the testing process.

\section{Experiment}

\texttt{\textit{SVIP-Reward}} is a reward model that can enhance MLLM or be used independently. To demonstrate its robustness and superiority, we outline our experimental setup (Section~\ref{sec5.1}). Then present three key experiments showing that the \texttt{\textit{SVIP-Reward}} can: 1) achieve SOTA performance on \textit{SVIP-Test} and other benchmarks (Section~\ref{sec5.2}). 2) provide reward signals during the reinforcement learning phase to improve the reasoning capabilities of MLLM  (Section~\ref{sec5.3}); 3) evaluate model outputs during inference to find the optimal solution (Section~\ref{sec5.4});  Finally, extensive ablation studies are conducted to further assess the contributions of different components of our approach (Section~\ref{sec5.5}).

\subsection{Experimental Setup}
\label{sec5.1}

\textbf{Model Setup.} We trained \texttt{\textit{SVIP-Reward}} on two backbones with different parameters: Qwen2-VL-7B~\cite{wang2024qwen2vlenhancingvisionlanguagemodels} and InternVL-2.5-2B~\cite{chen2025expandingperformanceboundariesopensource}. During label generation, we used GPT-4o mini~\cite{gpt4o} via API calls, which can accept multimodal inputs, serving as both the code generator and the model for generating CoT Steps. We set the number of steps in the least-to-most generation to two ($X=2$). Besides, all other model configurations were aligned with those used in PropTest~\cite{koo2024proptest}, De-fine~\cite{gao2024fine}, and ViperGPT~\cite{Suris_2023_ICCV}.

\textbf{Reward Model Settings.} The effectiveness of \texttt{\textit{SVIP}} is demonstrated through its reward model. To fair comparison, we employed three different experimental settings:
\begin{itemize}
\item \textit{Zero-shot}: This approach directly prompts the backbone model in zero-shot to derive multi-dimensional scores for CoT steps.

\item \textit{Tuning}: We fine-tune the backbone model using \textit{SVIP-Train} data for CoT tuning, with the same scoring head to predict the reward score.

\item \texttt{\textit{SVIP-Reward}}: Training the backbone by \textit{\textbf{TriAtt-CoT}} to predict the the reward score.
\end{itemize}

\textbf{Baseline.} We introduce additional baselines for comparison, including GPT-4o~\cite{gpt4o}, DeepSeek-VL-7B~\cite{lu2024deepseek}, Openflamingo-3B~\cite{awadalla2023openflamingo}, InternLM-XComposer-2.5-Reward-7B~\cite{zang2025internlm}, to further evaluate the performance of our method.

\textbf{Benchmark.} We extended our evaluation to benchmarks such as  MME~\cite{fu2024mmecomprehensiveevaluationbenchmark}, MMMU~\cite{yue2024mmmu}, MMMU-Pro~\cite{yue2024mmmuprorobustmultidisciplinemultimodal}, MathVista~\cite{lu2024mathvistaevaluatingmathematicalreasoning}, MMT~\cite{ying2024mmtbenchcomprehensivemultimodalbenchmark}, and POPE~\cite{li2023evaluating}. These benchmarks are specifically designed to assess the model’s capabilities in multi-modal understanding and hallucinations.

\begin{table*}
\centering 
\caption{Comparison of various benchmarks in training phrase.}
\label{table3}
\resizebox{\textwidth}{!}{
\begin{tabular}{l | l | c c c c c c} 
\midrule
 MLLM & Reward Model & MME$_{sum}$ & MMMU$_{val}$ & MMMU-Pro$_{overall}$ & MathVista$_{mini}$ & MMT$_{val}$ & POPE$_{avg}$ \\
\midrule
DeepSeek-VL-7B & - & 1835 & 36.9 & 18.2 & 36.7 & 53.6 & 87.9 \\
\midrule
Qwen2-VL-7B & - & 2374 & 58.5 & 30.9 & 59.3 & 64.6 & 89.3 \\
Qwen2-VL-7B & Qwen2-VL-7B (\textit{Zero-shot}) & 2390 & 59.9 & 31.3 & 59.4 & 64.8 & 89.5\\ 
Qwen2-VL-7B & Qwen2-VL-7B (\textit{Tuning}) & 2387 & 60.7 & 32.4 & 59.6 & 67.4 & 89.6 \\ 
Qwen2-VL-7B & Qwen2-VL-7B (\texttt{\textit{SVIP-Reward}}) & \textbf{2466} & \textbf{61.8} & \textbf{33.0} & \textbf{60.6} & \textbf{70.9} & \textbf{90.1}\\
\midrule
Openflamingo-3B & - &  912 & 37.2 & 22.0 & 33.2 & 51.4 & 73.4\\
\midrule
InternVL-2.5-2B & - & 2144 & 59.4 & 24.3 & 51.7 & 55.8 & 90.6\\
InternVL-2.5-2B & InternVL-2.5-2B (\textit{Zero-shot}) & 2202 & 60.8 & 24.7 & 52.5 & 56.2 & 90.6\\
InternVL-2.5-2B & InternVL-2.5-2B (\textit{Tuning}) & 2261 & 61.2 & 26.1 & 53.6 & 58.6 & 90.8\\ 
InternVL-2.5-2B & InternVL-2.5-2B (\texttt{\textit{SVIP-Reward}}) & \textbf{2332} & \textbf{62.0} & \textbf{27.4} & \textbf{54.9} & \textbf{63.5} & \textbf{90.8}\\
\midrule
\end{tabular}
}
\end{table*}

\begin{table*}
\centering 
\caption{Comparison of various benchmarks in inference phrase.}
\label{table4}
\resizebox{\textwidth}{!}{
\begin{tabular}{l | l | c c c c c c} 
\midrule
 MLLM & Reward Model & MME$_{sum}$ & MMMU$_{val}$ & MMMU-Pro$_{overall}$ & MathVista$_{mini}$ & MMT$_{val}$ & POPE$_{avg}$ \\
\midrule
GPT-4o & -  & 2328 & 69.1 &  51.9 & 63.8 & 65.4 & 86.9  \\
Deepseek-VL-7B & - & 1847 & 36.6 & 18.1 & 36.1 & 53.2 & 88.1 \\
Openflamingo-3B & - & 668 & 21.8 & 11.6 & 29.5 & 47.4 & 58.6 \\
\midrule
Qwen2-VL-7B & - &  2327 & 54.1 & 30.5 & 58.2 & 64.0 & 88.1\\
Qwen2-VL-7B & Qwen2-VL-7B (\textit{Zero-shot}) & 2301 & 59.1 & 30.7 & 59.0 & 64.8 & 89.0\\ 
Qwen2-VL-7B & Qwen2-VL-7B (\textit{Tuning}) & 2383 & 60.4 & 31.4 & 59.2 & 67.7 & 89.5\\ 
Qwen2-VL-7B & Qwen2-VL-7B (\texttt{\textit{SVIP-Reward}}) & \textbf{2472}& \textbf{61.3} & \textbf{32.3} & \textbf{61.6} & \textbf{70.5} & \textbf{90.2}\\
\midrule
InternVL-2.5-2B & - & 2138 & 40.9 & 23.7 & 51.3 & 54.5 & 90.6\\
InternVL-2.5-2B & InternVL-2.5-2B (\textit{Zero-shot}) & 2185  & 41.2 & 24.4 & 53.1 & 56.1 & 90.4\\
InternVL-2.5-2B & InternVL-2.5-2B (\textit{Tuning}) & 2246 & 43.0 & 25.9 & 53.7 & 58.6 & 90.8 \\ 
InternVL-2.5-2B & InternVL-2.5-2B (\texttt{\textit{SVIP-Reward}}) & \textbf{2319} & \textbf{43.7} & \textbf{26.8} & \textbf{55.4} & \textbf{63.3} & \textbf{90.9}\\
\midrule
\end{tabular}
}
\end{table*}

\begin{figure}
  \centering
    \includegraphics[width=\linewidth]{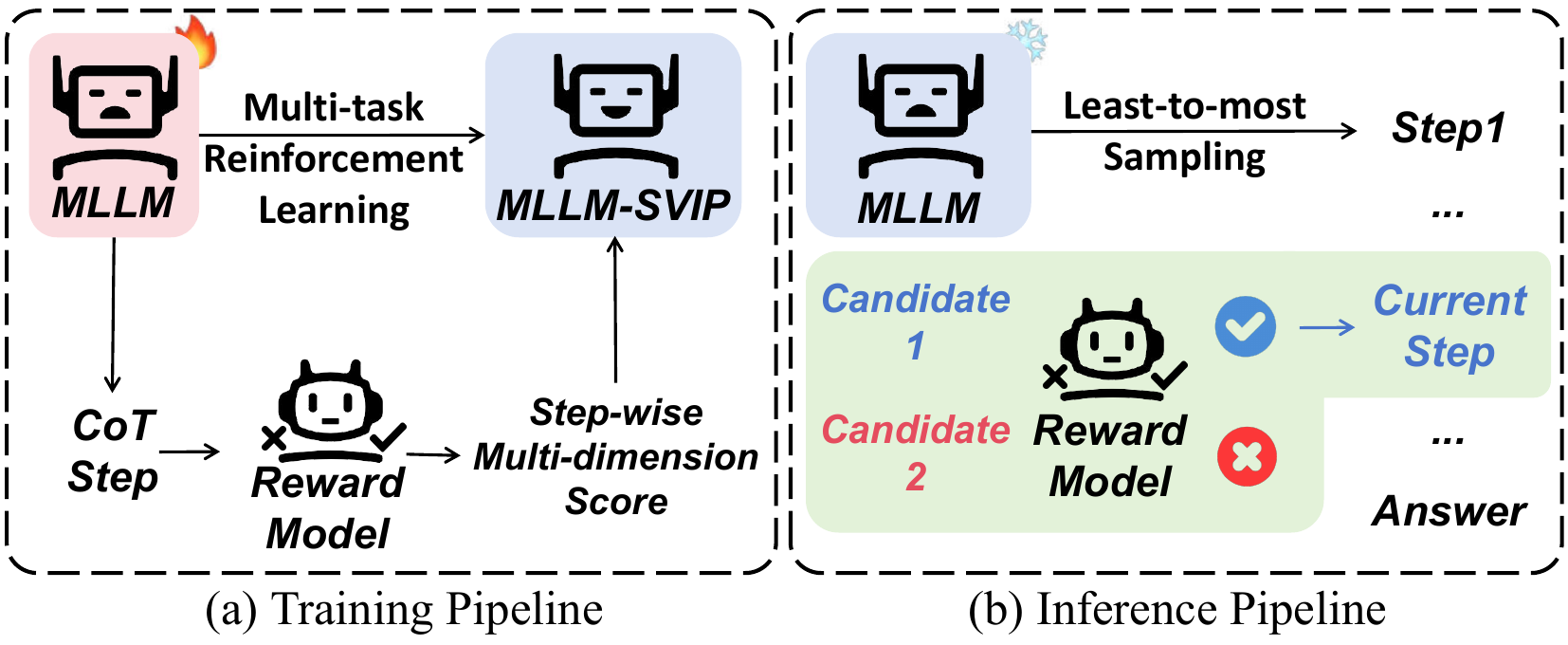}
    \vspace{-7mm}
    \caption{An illustration of reward models usage.}
    \label{fig_5}
    \vspace{-5mm}
\end{figure}

\subsection{\textbf{\texttt{\textit{SVIP-Reward}}} ~Test on CoT Benchmark}
\label{sec5.2}

We chose to evaluate the ability of the CoT reward model to accurately assess the three-dimensional labels of the given CoT steps on the \textit{SVIP-Test} and other benchmarks. The experimental results are shown in Table~\ref{table2}.

Most models, which have not undergone step-level training, performed poorly on \textit{SVIP-Test}, despite achieving good results on other benchmarks. This highlights that these models are trained with final outcomes as the sole learning objective, neglecting the impact of intermediate-step quality. It also reveals the limitations of benchmarks that focus only on the final results. In contrast, due to \textit{\textbf{TriAtt-CoT}} precisely differentiate between various dimensions, \textit{\texttt{SVIP}} demonstrates superior performance compared to models trained on the same data. This offers a robust paradigm for both the training and testing of the CoT reward model.

\subsection{\textbf{\texttt{\textit{SVIP-Reward}}} ~for RL Training}
\label{sec5.3}

\texttt{\textit{SVIP-Reward}} provides fine-grained reward signals during training to enhance MLLMs' reasoning capabilities. We used Qwen2-VL-7B~\cite{wang2024qwen2vlenhancingvisionlanguagemodels} and InternVL-2.5-2B~\cite{chen2025expandingperformanceboundariesopensource} as base models, employing three reward models as outlined in the reward model settings for multi-task reinforcement learning. The training pipeline is shown in Figure~\ref{fig_5}(a). To avoid overlap with \textit{SVIP-Train} or \textit{SVIP-Test} benchmarks, we used a mixture of the training dataset (as detailed in Appendix B), with results presented in Table~\ref{table3}.

The results show that models using the reward model outperform those without, achieving higher scores than the baselines. This improvement stems from the fact that instruction tuning alone cannot provide multi-dimensional reward scores. The \textit{Tuning} reward model outperforms the \textit{Zero-shot} model due to its explicit training in multi-step reasoning, enhancing its ability to handle complex, multi-step tasks. However, \texttt{\textit{SVIP-Reward}} surpasses \textit{Tuning} by employing a training method more suited for multi-label learning, making it a specialized model for evaluating CoT steps. Notably, InternVL-2.5-2B \texttt{\textit{SVIP-Reward}} achieves a higher score than both \textit{Zero-shot} and \textit{Tuning} methods on the MMMU-Pro benchmark by 2.7\% and 1.3\%. This plug-and-play reward model can be conveniently adapted to various reinforcement learning paradigms, significantly reducing the overhead associated with Reinforcement Learning from Human Feedback (RLHF).

\subsection{\textbf{\texttt{\textit{SVIP-Reward}}} ~for Inference-Time Scaling}
\label{sec5.4}

Inference-time scaling improves task accuracy by evaluating output candidates during the inference phase. Encouragingly, \texttt{\textit{SVIP-Reward}} is born to facilitate the evaluation of MLLM outputs during inference to identify the optimal solution. In the inference phase, we use the least-to-most method to let MLLM generate $N=4$ candidates, and \texttt{\textit{SVIP-Reward}} evaluates the CoT steps to select the best result (see Appendix C for details). The inference pipeline is shown in Figure~\ref{fig_5}b, with results in Table~\ref{table4}.

The results demonstrate that, compared to directly output answers, the inference-time scaling with a reward model, significantly enhances the model's reasoning capabilities. Although the reward model trained via \textit{Tuning} is better equipped to handle the increasing complexity of tasks, outperforming \textit{Zero-shot} models that struggle with scalability due to the lack of structured multi-step reasoning, it still falls short of \texttt{\textit{SVIP-Reward}}. This highlights that when filtering and ranking candidate steps, \texttt{\textit{SVIP-Reward}} delivers evaluations that are more accurate and reliable than other settings.

\textbf{Analysis on inference time scaling step candidate.} We further demonstrate that varying the number of next steps $N$ in the least-to-most generation is influential for scaling the inference-time capabilities of InternVL-2.5-2B. As shown in Figure~\ref{fig_6b}, increasing the number of candidate steps generated at each stage indeed benefits the model by expanding the exploration paths. However, considering the balance between resource utilization and performance enhancement, we uniformly adopted $N=4$ in our experiments.

\subsection{Ablation Experiment}
\label{sec5.5}

\begin{table}
\centering 
\caption{Ablation study on individual components.}
\vspace{-3mm}
\label{table5}
\resizebox{0.95\columnwidth}{!}{
        \begin{tabular}{ l | l | c  c } \midrule
        & & VLRewardBench & \textit{SVIP-Test} \\ \midrule
        0 & \makecell[l]{Qwen2-VL-7B \\ \texttt{\textit{SVIP-Reward}}} & 68.0 & 70.7 \\  \midrule
        1 & w/o \textit{TriAtt-CoT}  & 55.4  &  67.2 \\ 
        2 & w/o multi-dimensional label  & 55.2 &  66.8 \\ 
        3 & w/o CoT conversion  & 51.3 &  63.1 \\ 
        4 & w/o visual program  & 42.9 &  61.8 \\ \midrule
        \end{tabular}
       }
\vspace{-5mm}
\end{table}

\textbf{Analysis on \texttt{\textit{SVIP}} method.} To assess the impact of each module in generating step-level, multi-dimensional data within \texttt{\textit{SVIP}}, we designed an ablation study detailed in Table~\ref{table4}. 0) \texttt{\textit{SVIP-Reward}}: Visual programming-generated code blocks are converted into natural language with defined three-dimensional labels; 1)w/o TriAtt-CoT: Only use a linear layer after the model to get reward scores. 2) w/o multi-dimensional label: Labels are combined using an AND operation into a single reward signal; 3) w/o CoT conversion: Code blocks remain unconverted, and the CoT step involves only concatenating code and execution variables; 4) w/o visual program: Relies solely on the zero-shot backbone to decompose tasks and assign scores.

Based on the result analysis compared to \texttt{\textit{SVIP}}, we conclude that: 1)The lack of a multi-head attention structure would lead to the conflation of the three-dimensional labels, thereby reducing the method to a unidimensional performance. 2) Without multi-dimensional reward signals, the model struggles with complex and dynamic tasks due to suboptimal behavior optimization. 3) Models lacking code pre-training face challenges when using Python code and execution trajectories as CoT steps. 4) Without the cross-validation provided by code, the labels become unreliable, thereby diminishing the performance of the reward models.

\begin{figure}[!t]
\centering
\subfloat[]{
        \label{fig_6a}
		\includegraphics[scale=0.2]{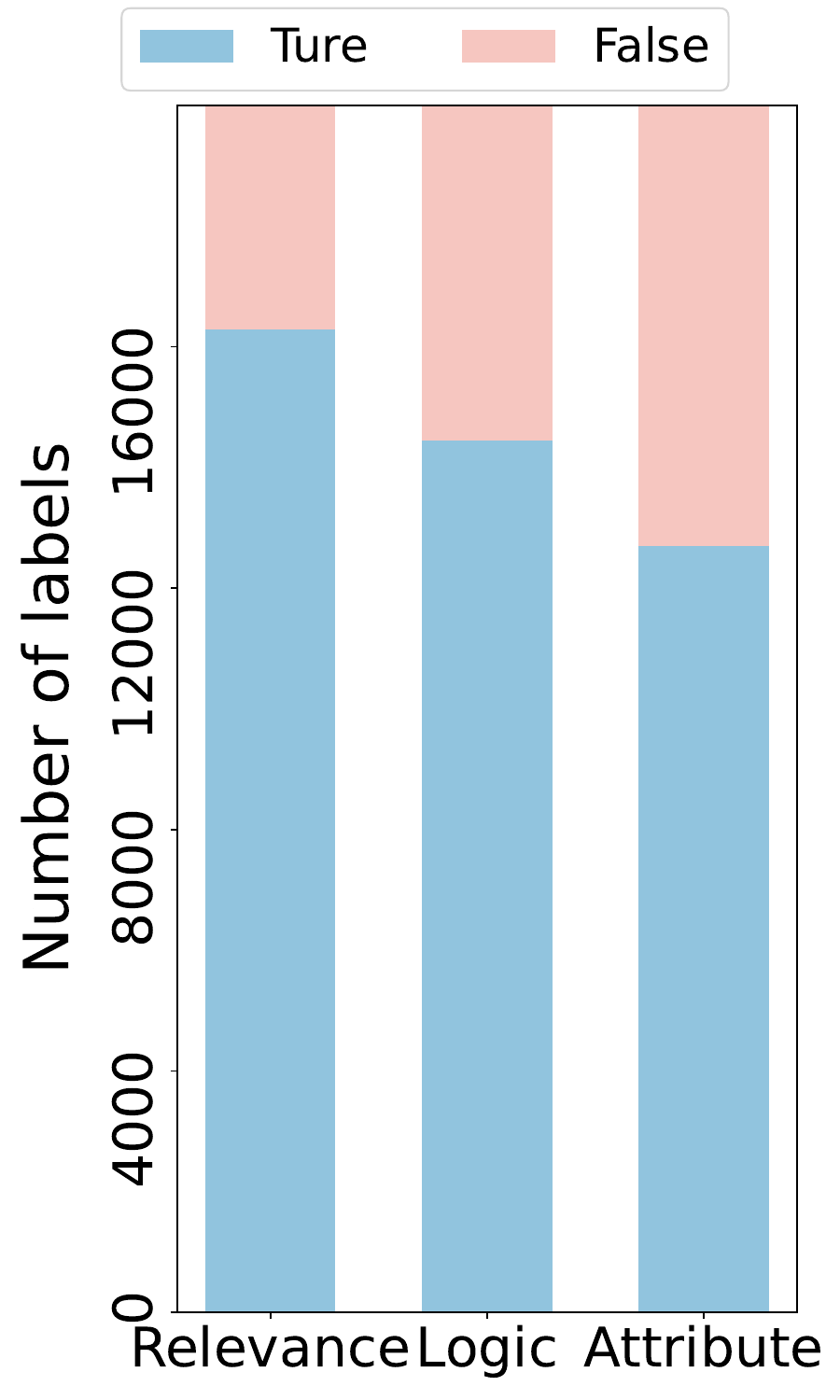}}
\subfloat[]{
        \label{fig_6b}
		\includegraphics[scale=0.4]{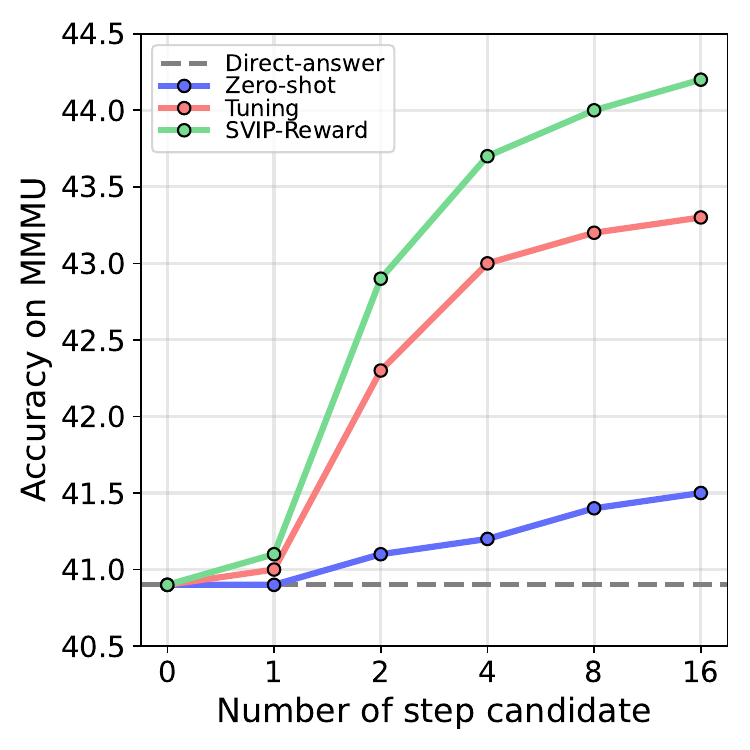}}
\vspace{-3mm}
\caption{Analysis on (a) label distribution of \textit{SVIP-Train} and (b) the number of candidates in inference scaling.}
\label{fig_6}
\vspace{-3mm}
\end{figure}

\textbf{The impact of multi-dimensional labels.} Based on the analysis of the number of SVIP-Train labels depicted in Figure~\ref{fig_6a} and the results presented in Table~\ref{table6}, our ablation studies elucidate the impact of multi-dimensional labels on the model. Among these labels, the Logic labels exert the most significant influence on model performance. This can be attributed to the fact that Logic labels encapsulate essential relationships within the reasoning process. Without explicit supervisory signals from these Logic labels, the reward model struggles to coherently guide the model through the reasoning steps, resulting in diminished accuracy.

\begin{table}
\centering
\caption{Ablation results of multi-dimensional labels.}
\vspace{-3mm}
\label{table6}
\resizebox{\columnwidth}{!}{
\begin{tabular}{clccccc}
\midrule
& & \multicolumn{3}{c}{Label} & \multicolumn{2}{c}{Accuracy} \\ 
\cmidrule(lr){3-5}  \cmidrule(lr){6-7} 
& & R & L & A & VLRewardBench & \textit{SVIP-Test} \\  
\midrule
0&Backbone & & &  & 35.1 & 59.6 \\
\midrule
1&\quad+ R & \checkmark &  &  & 44.9 & 60.3 \\
2&\quad+ L & & \checkmark &  & 46.5 & 61.8 \\
3&\quad+ A & & & \checkmark  & 43.7 & 60.5 \\
\midrule
4&\quad+ R + L & \checkmark & \checkmark &  & 51.8 &  64.0\\
5&\quad+ R + A & \checkmark & & \checkmark  & 49.6 &  63.4 \\
6&\quad+ L + A & & \checkmark & \checkmark  & 52.1 & 64.9 \\
\midrule
7&\makecell[l]{InternVL-2.5-2B\\ \texttt{\textit{SVIP-Reward}}} & \checkmark & \checkmark & \checkmark  & 56.3 & 66.5 \\
\midrule
\end{tabular}
}
\end{table}

\section{Conclusion}

We introduced \textit{\texttt{SVIP}}, a visual programming based approach to automatically train a step-level, multi-dimensional CoT reward model. By converting code generation and analysis into CoT statements, \textit{\texttt{SVIP}} facilitates detailed evaluations at each step, enhancing data cleaning, reinforcement learning, and test-time scaling. This approach not only improves MLLM performance but also establishes a comprehensive benchmark for training and testing reward models. \textit{\texttt{SVIP}} significantly increases the transparency and interpretability of reward mechanisms.

\clearpage

{\small
\bibliographystyle{ieeenat_fullname}
\bibliography{main}
}

\appendix
\onecolumn

\section{Case study}

\noindent 1) \textbf{Query: What is the name of the landmark in the picture?} 

Golden answer: Trondheimsfjorden

CoT step: \textit{We attempt to analyze information about a landmark in an image, but I don't know what island it is.}

Relevance: \textcolor{blue}{True}; Logic: \textcolor{blue}{True}; Attribute: \textcolor{blue}{False}.

\begin{figure}[H]
    \centering
    \includegraphics[width=0.5\linewidth]{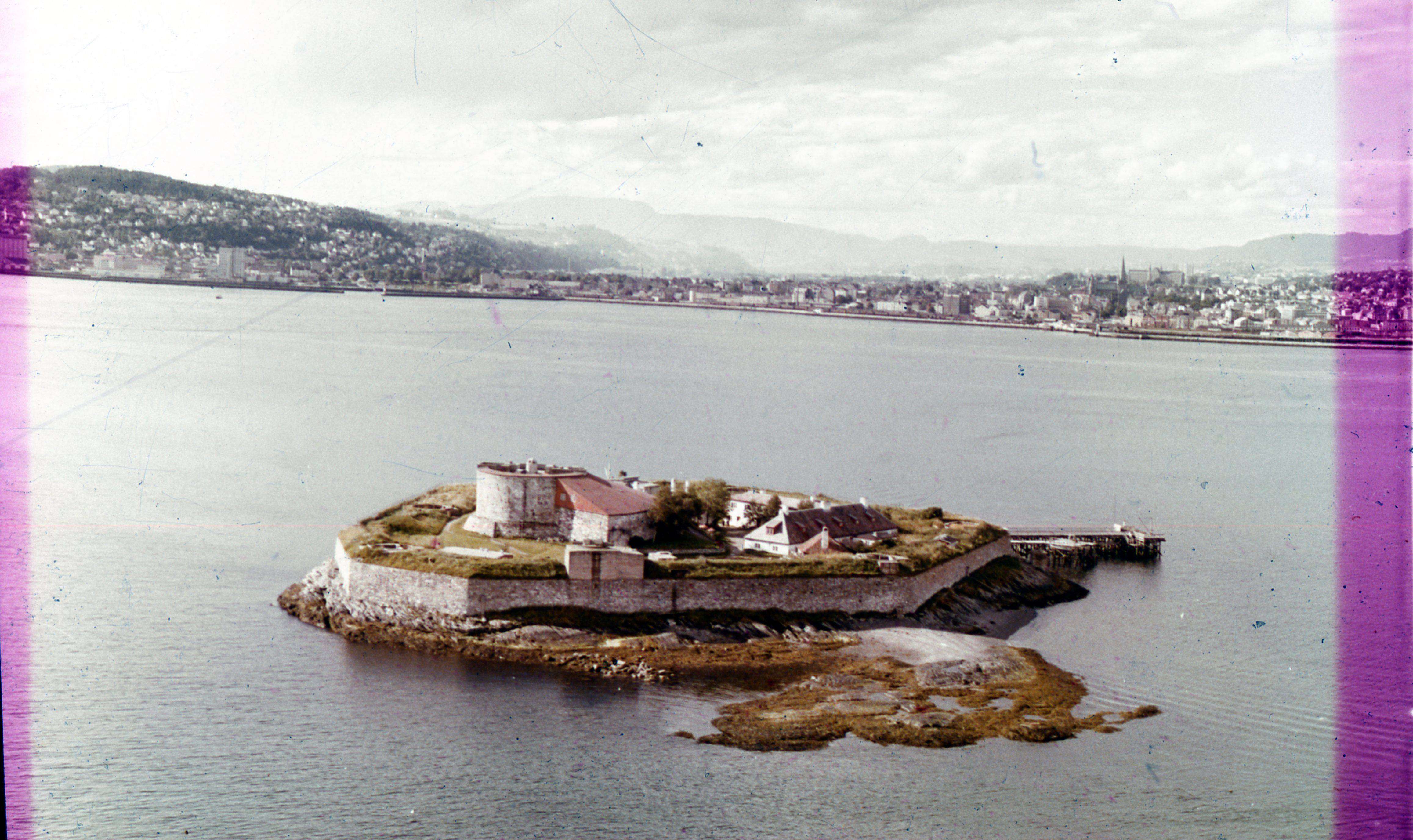}
    \caption{Attribute errors due to external knowledge errors.}
    \label{fig_appendix1}
\end{figure}

In this scenario, if GPT4V is utilized as the visual module for visual programming, it would accurately identify the landmark in the image as Munkholmen, a small islet located in the Trondheim Fjord in Norway. However, less capable models, such as blip2, would fail to provide the correct answer, subsequently leading to their detection by the SVIP system, which would mark the Attribute as false.

\noindent\textbf{Failure case:}

\noindent 1) \textbf{Query: What is the length of the sides in the triangle below?} 

Golden answer: 10

CoT step: \textit{Finally, we leverage the relationship between the area of an isosceles right triangle and its side lengths to determine the length of one side. Given an area of 200, we rearrange the area formula to express the side length in terms of the area. By calculating the square root of twice the area, we find that each side measures \textcolor{red}{20.0} units, confirming the triangle's dimensions.}

Relevance: \textcolor{blue}{True}; Logic: \textcolor{blue}{True}; Attribute: \textcolor{red}{True}.

\begin{figure}[H]
    \centering
    \includegraphics[width=0.35\linewidth]{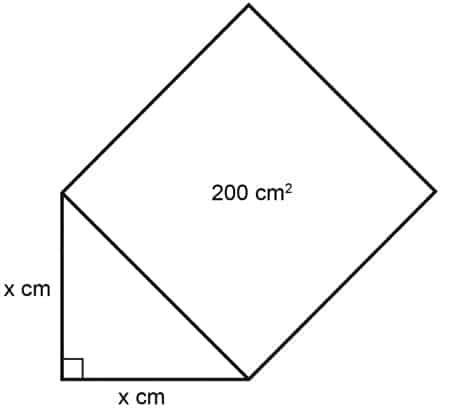}
    \caption{Enter Caption}
    \label{fig_appendix3}
\end{figure}

In this example, the response exhibits an attribute error, as calculations should show that the sides of the triangle should be 10. However, SVIP failed to accurately identify the erroneous attribute of 20.0.

\section{Details of training dataset}

The details of the data mixture of image caption and VQA datasets used in MLLM training are shown in Table \ref{table_appendix}. We only use a subset of each dataset’s training set. It should be noted that these data do not overlap with the data in Seedbench2. SVIP-Reward will provide step-level multi-dimensional reward signals for MLLM to complete these visual tasks during the reinforcement learning process.

\begin{table}[H]
\centering 
\caption{Data mixture of image caption and VQA datasets used in MLLM training.}
\label{table_appendix}
{
\begin{tabular}{l | l | r } \midrule
Dataset & Description & Number of samples \\ \midrule
COCO &  Scene description & 10.0K \\ 
Flickr 30K &  Scene description & 10.0K\\
VQAv2 & General &100.0K\\ 
GQA & Compositional & 86.0K \\ 
OK-VQA & Knowledge & 9.0K \\
TallyQA & Counting & 48.4K \\ 
\midrule
Total & & 263.4K\\
\midrule
\end{tabular}
}
\end{table}

\section{Candidate sorting rules during inference time scaling}

\begin{figure}[H]
  \centering
    \includegraphics[width=0.5\linewidth]{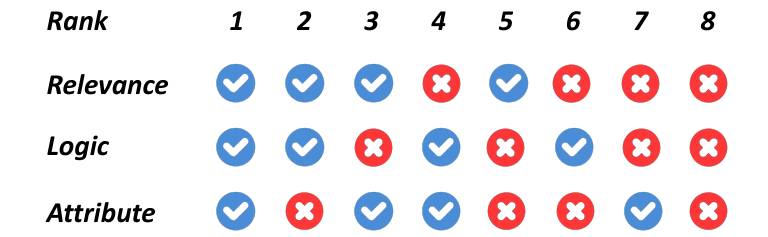}
    \caption{Candidate sorting rules.}
    \label{fig_appendix}
\end{figure}

During the inference phase, it is essential to select the most appropriate candidate from N options. Given that the output of the SVIP-Reward is not a continuous score, we sort the three existing labels and establish the following selection criteria for choosing steps: When comparing two candidate steps, the one with a higher count of correct labels is prioritized. Subsequently, they are ranked based on the order of relevance, logic, and attributes as key indicators. For candidates that have identical labels across all three dimensions, an MLLM is utilized to discern the superior option. This process is particularly manageable given that there are only eight possible scenarios during reasoning. So we also show the ranking in Figure~\ref{fig_appendix}.

During test-time scaling, there remains significant room for optimization in sampling and sorting, such as utilizing Monte Carlo Tree Search or Lookahead search strategies. Each method has its own corresponding and more efficient sorting approach. These aspects will be further explored in our dedicated research on test-time scaling.

\section{Prompt used in the method.}
We show the prompts used in the experiment as follows:
For the given problem, generate the code for the first step:

\begin{tcolorbox}[title = {Least-to-most code generation for the first step}]
Question: [QUESTION] 

Notice: Determine the first step for implementing this question and output the corresponding code. In details, you MUST write a descriptive comment for the chosen first step firstly and then write the corresponding code for the first step. Ensure the comment format should be: `\# Step 1: [Description of the step]'. Remember, you are required to provide only the descriptive comment and the corresponding codes for only the first step. Do not include any function definitions or return statement.
\end{tcolorbox}

\begin{tcolorbox}[title = {Least-to-most code generation for the following step}]
Question: [QUESTION]

Below are a few lines of code intended to solve the problem, which includes all the steps completed so far. If you think the existing code is sufficient to solve this question, make sure the code first prints the final result (which can be represented by a variable in the code) and then outputs `Work is Done!'. Otherwise, you should determine the next appropriate step and its corresponding code based on the context of the existing code. In details, you only need to output a descriptive comment for your current chosen step and then write the corresponding codes for that step, and do not include any previous steps, function definitions or return statement. Ensure the comment format is: `\# Step $N$: [Description of the step]', replacing $N$ with the current step number.

[All the code steps completed so far]
\end{tcolorbox}

\begin{tcolorbox}[title = {Convert code snippets to corresponding step CoT}]
Task Description:

Please generate a single-step Chain-of-Thought (CoT) for the current step based on the provided code block and the values of the intermediate variables.

Requirements:

- The Chain-of-Thought should begin with ``In this step, we use...".

- The explanation should not directly reference the variable names from the code block, but should include descriptions of the intermediate variables and its values.

- The Chain-of-Thought should be a concise, single-step explanation describing the logic of the code and how the intermediate variables support the reasoning.

- The description should be accurate, clear, and aligned with the programming logic. 

Code Block: [CODE\_BLOCK]

Intermediate Variables: [Values of intermediate variables]
\end{tcolorbox}

\begin{tcolorbox}[title = {PropTest's prompt for getting logic label}]
\# CONTEXT \#

The `solve\_query' function is a Python function that takes an image as input and implements the functionality described in [QUERY].

\# OBJECTIVE \#

Develop a Python function named `execute\_test' to verify whether the `solve\_query' function correctly implements the functionality described in [QUERY].

[EXAMPLES] are the in-context examples.

Include up to four test cases, each with the comment `\# Test case n:' above the assert statement, starting from 1.

Consider these guidelines when creating the test cases:

1. Keep in mind that the return values do not contain numbers.

2. If the Query is True or False questions, the return values will be yes or no.

3. If the Query gives options using``or", the return values will be one of the options.

4. Use the llm\_query function to answer informational questions not concerning the image.

\# STYLE \#

technical, in a correct Python format

\# TONE \#

clear, precise, professional

\# AUDIENCE \#

Developers and engineers who will use the test functions to verify the correctness of the solve\_query function

\# RESPONSE \#

Provide the function that start with`def execute\_test(image)' without any explanation.

Each test case should be commented with `\#Test case n:' where `n' represents the test case number.

\#\#\#
Here are some [EXAMPLES]: [EXAMPLES]
\#\#\#

\# Instruction \#

Generate the function execute\_test for the following query:

[Query]: INSERT\_QUERY\_HERE
\end{tcolorbox}

\begin{tcolorbox}[title = {Define's prompt for getting attribute label}]

\textbf{Visual Feedback (Image caption)}:

Please give me a short caption of [IMAGE2] located in the [BBOX] of [IMAGE1]. 

It mainly focuses on the connection between the object itself and the overall picture:

\textbf{Visual Feedback (Sub-step verification)}:

[IMAGE\_LIST] Is it possible that these pictures are the output of [substeps]?

\textbf{Textual Feedback (Text summarization)}:

[TRACE] Please summarize in natural language what program trace outputs.

\textbf{Textual Feedback (Logical verification)}:

[SUB\_STEPS] Is it possible that these sub\_steps can solve [QUERY]? 

If not, please modify it and only start with \#step1:, \#step2 etc.:

\textbf{Evaluator Prompt}:

You are a program evaluator.

I will provide you with a piece of code and the feedback and results after it is executed.

Your task is to compare the target of the code and its result after calling the external function.

[Query]: [QUERY]

[Code]: [ORIGIN\_CODE] 

[Intermediate Variables]: [Values of intermediate variables]

[Visual Feedback]: [FEEDBACK\_V]

[Textal Feedback]: [FEEDBACK\_T]

[Compile Feedback]: [FEEDBACK\_C]

Evaluate whether the output of the current code is correct:

\end{tcolorbox}

\end{document}